# Continuation Methods for Mixing Heterogeneous Sources

**Adrian Corduneanu**
Artificial Intellingence Laboratory
Massachussetts Institute of Technology
Cambridge, MA 02139

**Tommi Jaakkola**
Artificial Intellingence Laboratory
Massachussetts Institute of Technology
Cambridge, MA 02139

## Abstract

A number of modern learning tasks involve estimation from heterogeneous information sources. This includes classification with labeled and unlabeled data as well as other problems with analogous structure such as competitive (game theoretic) problems. The associated estimation problems can be typically reduced to solving a set of fixed point equations (consistency conditions). We introduce a general method for combining a preferred information source with another in this setting by evolving continuous paths of fixed points at intermediate allocations. We explicitly identify critical points along the unique paths to either increase the stability of estimation or to ensure a significant departure from the initial source. The homotopy continuation approach is guaranteed to terminate at the second source, and involves no combinatorial effort. We illustrate the power of these ideas both in classification tasks with labeled and unlabeled data, as well as in the context of a competitive (min-max) formulation of DNA sequence motif discovery.

## 1 INTRODUCTION

Many important learning problems involve a compromise between competing sources of information. For example, in semi-supervised classification the available training examples typically include a limited set of labeled data as well as a large dataset of unlabeled examples. A number of other problems share the same abstract structure. For example, in estimating graphical models we often need to determine the equivalent sample size, i.e., how the prior model should count relative to the available data. While such "allocation" problems are ubiquitous, it is not clear how they should be solved in general. In other words, what principle we should use to determine how one privileged source (labeled set, prior) should be balanced relative to the other source(s) (unlabeled set, observed data).

The allocation problems are not stable in the sense that small changes in the allocation can result in drastic changes in the model. Standard algorithms such as EM for combining complete and incomplete data do not address (or are aware of) the problem.

We introduce a general algorithm based on globally convergent homotopy continuation (Chow, Mallet-Paret, and Yorke, 1978) that combines a preferential data source with another by following a continuous path of stationary points of the optimization criterion for possible weightings of the sources. The path of fixed points found by the algorithm is unique and has strong theoretical existence guarantees, and provides a rich basis for analyzing problems that feature a competition between potentially conflicting sources. We demonstrate, in particular, how continuation methods can be used to identify critical allocations of the two sources, i.e., allocations around which dramatic changes can take place with minimal changes in allocation. Such critical events can be exploited either to guarantee stability of estimation, or to ensure a substantial deviation from the initial data source.

We begin by introducing homotopy continuation in a classification setting and subsequently present relevant theoretical foundation. We extend the methodology in a number of relevant directions. Specifically, we provide results in text classification with naïve Bayes as well as solve the problem of finding DNA binding motifs in a competitive manner with homotopy continuation.

## 2 CLASSIFICATION WITH LABELED AND UNLABELED DATA

We commence here with the problem of improving classification accuracy by means of incorporating information from unlabeled examples. Unlabeled examples are typically easy to come by and often define the context in which the classification task has to be solved. Moreover, they provide information about the density and arrangement (clus-tering) of the examples. We restrict ourselves here to generative models but the methodology remains applicable to a wider class of approaches including margin based discriminative methods.

## 2.1 ESTIMATION AS SOURCE ALLOCATION

The objective here is to estimate the parameters of a joint distribution $p(z|\theta)$ from both labeled and unlabeled samples. Here $z = (x, y)$, where $y$ is the class label of $x$. For clarity, we define "incompleteness" of examples simply in terms of missing labels; this restriction is not necessary (cf. (Corduneanu, 2002)).

We assume that the joint distribution can be written as an (extended) exponential family where $\mathbf{t}(z)$ denotes the vector of sufficient statistics and $\psi(\theta)$ is the partition function (cumulant generating function):

$$p(z|\theta) = \exp\left(\theta \mathbf{t}(z)^T + k(z) - \psi(\theta)\right) \quad (1)$$

The standard approach in this setting is to maximize the likelihood of both labeled and unlabeled data (Nigam et al., 2000) via optimization algorithms that handle missing data such as the EM algorithm (Dempster, Laird, and Rubin, 1977). The combined likelihood criterion can be expressed in terms of minimizing

$$(1 - \lambda) D(\hat{p}^C(z) \| p(z|\theta)) + \lambda D(\hat{p}^I(x) \| p_x(x|\theta)) \quad (2)$$

where $p_x$ is the $x$ marginal of $p$; $\hat{p}^C$ and $\hat{p}^I$ refer to the empirical estimates from complete and incomplete data, respectively. From the point of view of maximum likelihood estimation, $\lambda$ should be set to the fraction of incomplete samples. Fixing $\lambda$ in this manner has no sound justification, however. Indeed, with a disproportionately large number of incomplete examples, $\lambda \approx 1$, and the criterion effectively omits[1] the few labeled examples defining the classification task. The solution should be sought with other allocations $\lambda \in [0, 1]$.

Perhaps surprisingly, the estimation problem with varying allocation is inherently unstable. To better illustrate this, we provide here an alternative but equivalent formulation of the estimation criterion. Specifically, we minimize the distance to the incomplete data source (empirical estimate) subject to a constrain that we remain sufficiently close to the complete data solution. More precisely,

$$\min_{\theta} D(\hat{p}^I(x) \| p_X(x|\theta)) \text{ s.t. } D(\hat{p}^C(z) \| p(z|\theta)) \leq \beta(\lambda) \quad (3)$$

where $\beta(\lambda)$ is a monotonically increasing function of the allocation $\lambda$ and $\beta(0) = \min_\theta D(\hat{p}^C(z) \| p(z|\theta))$. The equivalence is shown in the Appendix (Proposition 4).

Figure 1 illustrates why the estimation problem involves critical events (sudden drastic changes in the model). The objective minimizes the distance between a convex data manifold $\mathcal{D}$ (densities with marginal $\hat{p}^I$), and a non-convex marginal exponential family $\mathcal{M}$ (Csiszár and Tusnády, 1984), while bounding the distance from $\hat{p}^C$. As $\beta(\lambda)$ increases the bound becomes less restrictive, and the minimum may suddenly jump to a different convex region of $\mathcal{M}$. Another way of expressing the phenomenon is that while the achieved likelihood value changes smoothly with the allocation $\lambda$, the optimizing argument (the model) can incur drastic changes.

---
[1]Save perhaps in terms of initialization.

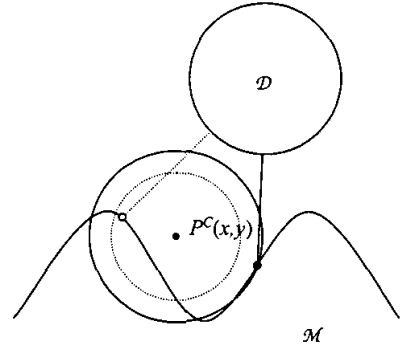

Figure 1: Minimum Distance May Jump as the Bound on Distance to $\hat{p}^C$ Loosens.

From the point of view of classification, it is crucial to understand when the critical events occur. The drastic changes at the critical points are after all no longer tied to the classification task (the few labeled examples). The critical points are, however, not visible within the EM algorithm and can be found only by tracing the solutions at increasing levels of allocation. The paths generated by evolving the solutions in this manner turn out to be unique (see later sections). They can also be found relatively efficiently.

## 2.2 PATH TRACING OF EM FIXED POINTS

We describe here the EM algorithm for classification with labeled and unlabeled data (cf. (Nigam et al., 2000)) and how the continuous paths of solutions can be traced in an efficient and stable manner by appeal to the theory of homotopy continuation.

With only unlabeled examples, the convergence points of the EM algorithm are fixed points of the EM iteration (operator): $\eta = \text{EM}_1(\eta)$, where

$$\text{EM}_1(\eta) = \frac{1}{M} \sum_{1 \leq j \leq M} E[\mathbf{t}((x_j^I, y))|\eta, x_j^I] \quad (4)$$

and the sum is over unlabeled samples $x_j^I$, and $\eta = E[\mathbf{t}(z)] = \frac{d}{d\theta}\psi(\theta)$ is the equivalent mean parameterization of the exponential family.

When we also have some labeled information, we can ad-just the EM operator so as to optimize Eq. (2):

$$\text{EM}_\lambda(\eta) = (1 - \lambda)\hat{\eta}^C + \lambda \text{EM}_1(\eta) \quad (5)$$

where $\hat{\eta}^C$ are the mean parameters of empirical $\hat{p}^C$. The key property here is the linear separation of labeled and unlabeled information in the mean parameterization.

Fixed points of $\text{EM}_\lambda$ range from the unique $\hat{\eta}^C$ at $\lambda = 0$ to multiple optima at $\lambda = 1$ which are indistinguishable with respect to label permutations (at least). Only one such permutation achieves good classification performance but the EM algorithm offers no guarantees finding it. In contrast, we identify only those fixed points that are firmly rooted in the labeled evidence by extending a continuous path of solutions $\text{EM}_\lambda(\eta) = \eta$, from $\lambda = 0$ up to the *critical allocation* $\lambda^*$.

A simple but numerically unstable method for tracing such paths is to solve the associated differential equation as a function of $\lambda$ (Corduneanu and Jaakkola, 2001); however, discontinuities arise from non-functional relations between the allocation $\lambda$ and the model (these are precisely the critical points) and this solution breaks down at/close to such points. An alternative approach is needed for stability and for going through the critical points.

Homotopy continuation eliminates the difficulty by following the path in the joint $(\eta, \lambda)$ space. The allocation $\lambda$ may now actually decrease along the path (upon hitting a critical point as defined previously). This is enough to remove discontinuities (which are almost surely not bifurcations) in all but measure zero cases (see Section 3). To evolve fixed points in the joint parameterization $(\eta, \lambda)$ we have to ensure that any adjustments to the solution do not violate the fixed point condition:

$$\left[\lambda \nabla_\eta \text{EM}_1(\eta) - I \quad \text{EM}_1(\eta) - \hat{\eta}^C\right] \cdot \begin{pmatrix} d\eta/ds \\ d\lambda/ds \end{pmatrix} = 0 \quad (6)$$

where $s$ may be chosen to be the unit length parameterization of the curve[2]. Provided that certain rank constraints discussed later in the paper are satisfied, the null space constraint above indeed defines a unique curve.

For exponential families the Jacobian $\nabla_\eta \text{EM}_1(\eta)$ of the EM operator has an explicit formula:

$$\nabla_\eta \text{EM}_1(\eta) = \left[\frac{1}{M} \sum_{1 \leq j \leq M} \text{cov}(\mathbf{t}(x_j^I, y)|\eta, x_j^I)\right] F(\eta)^{-1} \quad (7)$$

where cov is the covariance of the vector of sufficient statistics, and $F$ is the Fisher information matrix $\frac{\partial^2}{\partial \theta^2}\psi(\theta)$. (See (Corduneanu, 2002).)

---

[2]This is analogous to annealing procedures, where the state of the system may no longer be a function of temperature but rather a function of time.

Singularities of $\lambda \nabla_\eta \text{EM}_1(\eta) - I$ recover the discontinuities (critical points) where the naïve method breaks down. A stable identification of critical allocations, where the path looses connection with the initial data source, is a powerful feature of homotopy continuation. The critical allocations emerge as non-functional relations between the evolving model and the allocation $\lambda$ (see later in the paper). In settings other than classification, it may actually be desirable to follow through the critical points to ensure a substantial departure from the initial information source.

We present an algorithm based on Euler's method for path following, although in practice better ODE solvers are more appropriate (our implementation uses the Runge-Kutta method). See, e.g., (Watson and Morgan, 1987) or (Salinger et al., 2002) for a range of numerical algorithms for homotopy continuation. Starting at $\hat{\eta}^C$, we follow the path by finding the one-dimensional null space of the homotopy matrix in (6), and changing $(\eta, \lambda)$ in this direction. The orientation is ambiguous: we can move forward or backward along the unique path. We select the direction that makes the smallest angle with the previous path direction. In terms of complexity, the dominant step is solving (6) which involves $O(n^3)$ computation, where $n$ is the dimensionality of $\eta$.

## 3  CONTINUATION FROM A THEORETICAL PERSPECTIVE

### 3.1  GENERAL HOMOTOPIES

Continuation methods discussed above are based on the theory of globally convergent with probability one homotopy continuation, introduced in (Chow, Mallet-Paret, and Yorke, 1978). Watson gives a careful description of this method (Watson, 2000), and provides a numerical package HOMPACK for homotopy continuation (Watson and Morgan, 1987). We start by briefly elaborating the theoretical foundation.

Homotopy continuation is a general method for solving a nonlinear equation $g(x) = 0$, by starting with a trivial equation $f(x) = 0$ whose solution is known, and tracking its root while morphing $f$ into $g$. Formally, given $f, g : E^n \to E^n$, define a homotopy function to be a smooth map $h : E^n \times [0, 1] \to h^n$ such that $h(x, 0) = f(x)$ and $h(x, 1) = g(x)$. The goal becomes to find solutions of $h(x, \lambda) = 0$ for all $\lambda \in [0, 1]$. For instance, the nonlinear equation in classification with labeled/unlabeled data would be the fixed-point condition of the EM operator.

As discussed earlier, we can track solutions by solving the associated differential equation with respect to a path parameter $s$ and initial condition $f(x) = 0$ and $\lambda = 0$ at $s = 0$:

$$\nabla_{(x,\lambda)} h(x, \lambda) \cdot \begin{pmatrix} dx/ds \\ d\lambda/ds \end{pmatrix} = 0 \quad (8)$$

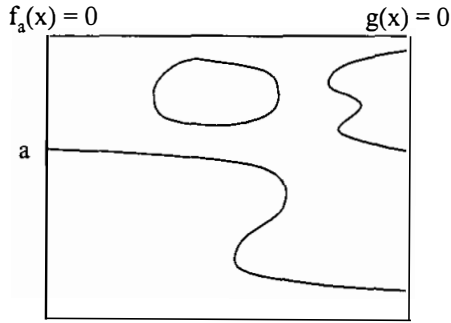

Figure 2: Possible Configurations of Roots of the Homotopy between $f_a(x)$ and $g(x)$

The above differential equation defines a unique path free of discontinuities and bifurcations as long as the homotopy matrix $\nabla_{(x,\lambda)} h(x, \lambda)$ has maximal rank along the path (cf. implicit function theorem). While this is not always the case, a simple change makes discontinuities very unlikely. The idea is to extend the analysis to a *family* of trivial equations $f_a(x) = 0$, with $a \in E^m$ an *initial point* that we are free to choose, that are all morphed into $g$ by the homotopy $h_a(x, \lambda) = h(a, x, \lambda)$. Then a continuous path of solutions from $f_a$ to $g$ exists for all but a measure 0 set of initial points $a$. This is a consequence of the following theorem proved in (Chow, Mallet-Paret, and Yorke, 1978) from *Parametrized Sard's Theorem*:

**Theorem 1** *Let $h : E^m \times E^n \times [0, 1) \to E^n$ be a $C^2$ map such that the $n \times (m + n + 1)$ Jacobian $\nabla_{(a,x,\lambda)} h$ has rank $n$ on the set $h^{-1}(0)$, and $h_a^{-1}(0)$ is bounded for fixed $a \in E^m$. Then $\nabla_{(x,\lambda)} h_a$ has full rank on $h_a^{-1}(0)$ except for possibly on a set of measure 0 of $a$'s.*

Thus when the hypothesis of the theorem holds and $a$ is not in the measure-zero set, any $(x, \lambda)$ root of $h_a$ can be continuously and uniquely extended in both directions of increasing and decreasing $s$. Assuming that $f_a$ has a unique root, if follows that all roots of $h_a$ must be on a unique continuous path of roots from $\lambda = 0$ to $\lambda = 1$, on closed continuous loops with $0 < \lambda < 1$, or on paths that start and end at $\lambda = 1$ (Figure 2).

Using homotopy continuation involves choosing an appropriate homotopy and set of initial points such that the first function has a unique root, and the hypothesis of the fundamental theorem holds.

The most difficult assumption to guarantee in Theorem 1 is that $\nabla_{(a,x,\lambda)} h$ has full rank. This rank constraint is trivially satisfied for (almost) all initial $a$'s under the following standard *fixed-point homotopy*:

$$h_a(x, \lambda) = (1 - \lambda)(a - x) + \lambda(g(x) - x) \qquad (9)$$

### 3.1.1 Fixed-point Homotopy for EM

The fixed points of the EM operator in the context of complete and incomplete information sources (5) can be seen as a standard *fixed-point homotopy* as defined in (9). To see this, simply replace $x$ with the mean parameters $\eta$, $\lambda$ remains the source allocation, and $\hat{\eta}^C$ plays the role of the initial point $a$.

In the case of EM, a necessary condition for the boundedness requirement of Theorem 1, needed to ensure that the path reaches $\lambda = 1$ in finite time, is that conditional expectations of sufficient statistics are always bounded. While this is not satisfied by all exponential families, most interesting cases will satisfy it, including all classification problems with finitely many labels. In addition, the EM operator is $C^2$ on valid mean parameters, thus the existence and uniqueness of a path of fixed-points from $\hat{\eta}^C$ to $\lambda = 1$ is assured.

Not all fixed points of the $EM_\lambda$ operator are valid[3] mean parameters, and one may wonder whether other constraints must be imposed along the path. Fortunately, the path started at $\hat{\eta}^C$ will automatically follow valid mean parameters for all $\lambda < 1$. If $C$ is the valid convex set of mean parameters, no fixed point with $\lambda < 1$ can be on the boundary of $C$, as a convex combination between $EM_1(\eta)$ on the closure of $C$, and the interior point $\hat{\eta}^C$. Thus the continuous path started at $\hat{\eta}^C$ cannot intersect the boundary of $C$, and it must be entirely contained in $C$.

## 3.2 LINEAR ALLOCATION OF OPTIMIZATION CRITERIA

We derive important properties valid along any continuous path of stationary points of a convex combination of optimization criteria. Let $F(x)$ and $G(x)$ be two optimization criteria, possibly derived from different sources of information, where $x$ are the parameters to be estimated. Let

$$J_\lambda(x) = (1 - \lambda)F(x) + \lambda G(x), \ \lambda \in [0, 1] \qquad (10)$$

be their convex combination, and consider a path of stationary points $\nabla_x J_\lambda(x) = 0$ in the $(x, \lambda)$ space that starts at $\lambda = 0$ and ends at $\lambda = 1$. We are interested in the monotonicity of the individual criteria $F(x)$ and $G(x)$ along the path.

A first result shows that such path cannot take the same $x$ value twice, unless it is a constant path. If $s$ is the path parameter, then:

**Theorem 2** *If $\frac{dx}{ds} = 0$ for some $s$, $\lambda(s) \in (0, 1)$, then $x(s)$ is constant for all $s$.*

---

[3]The operator itself does not explicitly guarantee that, e.g., certain mean parameters such as marginals are necessarily nonzero but rather maintains the property if it holds for the initial guess in the EM iterations.

**Proof** Any $x$ along the path satisfies $(1-\lambda)\nabla F(x) + \lambda\nabla G(x) = 0$, from which we obtain $(\nabla G(x) - \nabla F(x))\frac{d\lambda}{ds} + ((1-\lambda)\nabla^2 F(x) + \lambda\nabla^2 G(x))\frac{dx}{ds} = 0$. If $\frac{dx_0}{ds_0} = 0$ for some $x_0$, the previous equations yield $\nabla F(x_0) = \nabla G(x_0) = 0$, thus $(1-\lambda)F(x_0)+\lambda G(x_0) = 0$ for all $\lambda \in [0,1]$, and the path must be constant. □

More importantly, $F(x(s))$ and $G(x(s))$ can change their monotonicity only at critical points, and even then this is unlikely to happen. Thus typically $F$ always increases with $s$ while $G$ decreases, even if the path passes through critical allocations (assuming we are searching for minima). The property stems from the following formal result:

**Theorem 3** $\frac{d}{ds}F(x(s))$ and $\frac{d}{ds}G(x(s))$ can be 0 only when $\frac{d\lambda}{ds} = 0$, unless $x(s)$ is constant for all $s$.

**Proof** Assume $\nabla F(x)\frac{dx}{ds} = 0$ for some $s$ at which $\frac{d\lambda}{ds} \neq 0$. Then from $(\nabla G(x) - \nabla F(x))\frac{d\lambda}{ds} + ((1-\lambda)\nabla^2 F(x) + \lambda\nabla^2 G(x))\frac{dx}{ds} = 0$ we derive $(\nabla G(x) - \nabla F(x))\nabla F(x)^T = 0$. This combined with $(1-\lambda)\nabla F(x) + \lambda\nabla G(x) = 0$ yields $\nabla F(x) = \nabla G(x) = 0$, which again means that the same $x$ is a stationary point for all $\lambda$, and the path must be constant. □

For the monotonicity reversal of $F(x)$ to happen at a critical point, the gradient $\nabla F(x)$ must be orthogonal to $dx/ds$ exactly at the critical point, which rarely happens in practice.

Usually the starting criterion $F(x)$ is convex with a unique minimum, while $G(x)$ features many local minima. The continuous path of stationary points deterministically finds a preferential minimum of $G(x)$ reachable from the initial point by continuous descent in $G(x)$. Although monotonicity is not influenced by critical points, the minimum/maximum character of the stationary points must change after passage through critical allocation. Figure 3 illustrates this phenomenon in a one-dimensional problem.

## 4 TEXT CLASSIFICATION WITH NAÏVE BAYES

We illustrate homotopy continuation on text classification from labeled/unlabeled data on a discrete naïve Bayes model. We represent documents by word-appearance binary features, assumed to be independent given class label. Such assumption is tractable but inaccurate, and we expect homotopy continuation to be able to guard against model-misfit errors by limiting allocation below its critical value. We use standard naïve Bayes rather than a multinomial model (Nigam et al., 2000), for the purpose of introducing continuation on the simplest graphical model.

Observed data consists of features $(x_1, x_2, \ldots, x_k)$, and probabilities over documents satisfy $P(\mathbf{x}, y) = $

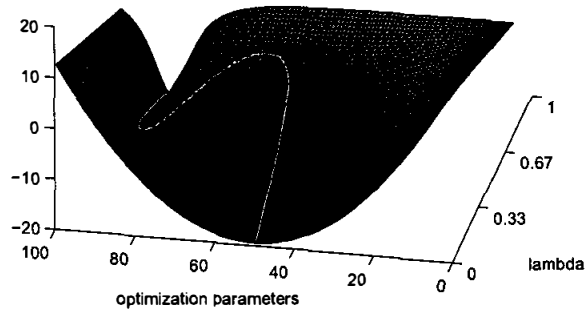

Figure 3: Continuous Path of Stationary Points Going through a Local Maxima Region between Critical Points

$\left[\prod_{i=1}^k P(x_i|y)\right] P(y)$. This defines an exponential family of mean parameters $P(y), P(x_i, y)$, and an $EM_\lambda$ operator

$$P(y) \leftarrow (1-\lambda)P^C(y) + \lambda\sum_{\mathbf{x}} P^I(\mathbf{x})P(y|\mathbf{x})$$
$$P(x_i, y) \leftarrow (1-\lambda)P^C(x_i, y) + \lambda\sum_{\mathbf{x}\setminus x_i} P^I(\mathbf{x})P(y|\mathbf{x})$$

where $P^C$ and $P^I$ are Laplace smoothed empirical frequencies from complete and incomplete samples.

In order to compute the EM Jacobian we only need derivatives of $P(y|\mathbf{x})$ with respect to mean parameters:

$$\frac{\partial P(y|\mathbf{x})}{\partial P(y')} = (k-1)\left[\frac{P(y|\mathbf{x})P(\mathbf{x}|y')}{P(\mathbf{x})} - \delta_{y'}(y)\frac{P(\mathbf{x}|y')}{P(\mathbf{x})}\right]$$

$$\frac{\partial P(y|\mathbf{x})}{\partial P(x_j, y')} = \delta_{x_j}(x_i)\left[-\frac{P(y|\mathbf{x})P(\mathbf{x}\setminus x_i|y')}{P(\mathbf{x})} + \right.$$
$$\left. + \delta_{y'}(y)\frac{P(\mathbf{x}\setminus x_i|y')}{P(\mathbf{x})}\right]$$

where $P(\mathbf{x}\setminus x_i|y')$ is marginalized over the $i$'th feature.

Then homotopy continuation proceeds as described in the general classification setting, stopping at critical allocation to preserve continuity with the complete or labeled information source.

### 4.1 TEXT CLASSIFICATION RESULTS

We use the *20-newsgroups* database for training (Nigam et al., 2000), which is a collection of newsgroup articles labeled by the newsgroup on which they were posted. To control the $O(n^3)$ complexity of the algorithm, we reduce the size of the problem to 3 classes, and 20 features selected by their mutual information with class labels. We perform feature selection only once on all 20 newsgroups, hoping that the most significant 20 words do not vary too much for the three-class problem with limited labeled samples.

Table 1: Error Rates of Maximum Likelihood from Labeled Data, Homotopy Continuation, and EM initialized based on the labeled estimate on 50 Random Selections of 10 Labeled Documents from 2926 Documents in Three Classes

|               | labeled only | homotopy | EM    |
|---------------|--------------|----------|-------|
| critical runs | 35.8%        | 20.4%    | 28.0% |
| all runs      | 35.7%        | 21.4%    | 27.7% |

In Table 4.1 we show results from an experiment with 10 randomly selected labeled documents, and the rest of 2916 available documents from the 3 classes selected (*talk.politics.mideast*, *soc.religion.christian*, and *sci.crypt*) as unlabeled data. The results combine 50 experiments out of which 45 featured homotopy paths with critical allocation. We break the results into an average over the 45 experiments with critical allocation, and an average over all runs. Maximum-likelihood allocation is effectively 1, while the average critical allocation was 0.93. We see that homotopy continuation dramatically improves the poor estimation based only on 10 labeled samples, and outperforms EM even if the critical allocation is close to the maximum likelihood allocation. This can happen essentially for two reasons: first, even a small increase in allocation can result in a jump followed by large increase in error rate, and, second, the EM algorithm, even with the same allocation, may converge to other fixed points no longer traceable to the initial labeled solution.

To illustrate the sudden degradation of performance after critical allocation we plot the evolution of error rate and allocation along the homotopy path (Figure 4). Note that different runs may produce different types of evolution, including no critical allocation, increasing error rate from the very beginning, optimal allocation less than the critical one, and even a sudden increase of performance at the critical allocation. However, critical allocation invariably signals a sudden change in performance, most often negative, showing that estimation beyond critical allocation is unstable.

## 5   COMPETITIVE MOTIF DISCOVERY

We provide here a novel competitive (minimax) formulation of DNA sequence motif discovery and show how the continuation approach developed earlier in the paper provides a natural solution.

In motif discovery we are given a set of DNA sequences corresponding to (upstream) intergenic regions (promoter regions) of selected genes. These sequences (each a few hundred bases long) potentially contain binding sites for transcriptional activators or other DNA binding proteins. Each binding site is characterized by a sequence of not necessarily adjacent bases whose identity is relatively conserved across sequences containing the same binding site or motif. The locations of the motifs along the sequences are unknown and each sequence may or may not contain any particular (unknown) motif.

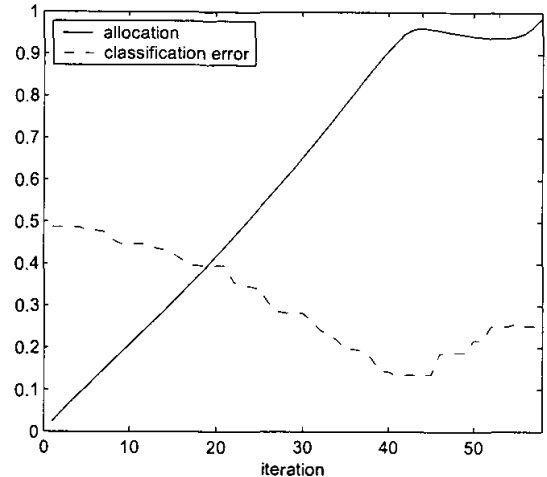

Figure 4: Evolution of Error Rate and Allocation with Homotopy Continuation Iteration on a Discrete Naïve Bayes Model with 20 Binary Features

Unlike previous approaches (Hughes et al., 2000), we formulate the problem as a direct competition between the motif model capturing the limited base conservation and the background model typically tailored to the overall frequence of bases. We seek motifs that are robust in the sense that they can distinguish themselves from a family of background models (specifically the worst case background).

Let $P_B(x)$ denote the background model over sequence $x$. For simplicity, we assume that the bases are independent, i.e., $P_B(x) = \prod_{t=0}^{L-1} B(x_t)$, where $L$ is the length of the sequence and $B(x_t)$ is multinomial over the four possible bases $\{A, G, T, C\}$. When $x$ contains a motif in position $t$, we replace the background predictions with those of $m$ (size of the motif) consecutive position specific multinomials. In other words,

$$P_{B,Q}(x|\text{motif in position } t) = P_B(x) \prod_{i=0}^{m-1} \frac{Q_i(x_{t+i})}{B(x_{t+i})} \quad (11)$$

When we add the uncertainty about the position and the existence of motif in $x$, we obtain the following likelihood ratio

$$\frac{P_{B,Q}(x)}{P_B(x)} = p^0 \sum_{t=0}^{L-m-1} p^0(t) \prod_{i=0}^{m-1} \frac{Q_i(x_{t+i})}{B(x_{t+i})} + 1 - p^0$$

where $p^0(t)$ is the (sequence specific) prior over positions (e.g., uniform) and $p^0$ is the prior probability that $x$ has the motif. The minimax problem we wish to solve is $\max_Q \min_B \sum_k \log P_{B,Q}(x^{(k)})/P_B(x^{(k)})$, where $k$ goes over the available sequences (minimax and maxmin are equivalent in this case).

In the continuation approach we trace a specific minimax solution of

$$(1 - \lambda)D(B^0\|B) - (1 - \lambda)\frac{1}{m}\sum_{i=0}^{m-1} D(B^0\|Q_i)$$
$$+\lambda \sum_k \log \frac{P_{B,Q}(x^{(k)})}{P_B(x^{(k)})} \qquad (12)$$

starting with $\lambda = 0$. Here $B^0$ is the reference background model and is based on the empirical frequence of bases in the available sequences. The fixed point equations resulting from the above objective are analogous to those discussed earlier (EM equations). Let $n_i(z) = \sum_k p_k p(t|k)\delta_{x_{t+i}}(z)$, where $p_k$ is the posterior probability of motif appearing in the $k^{th}$ sequence and $p(t|k)$ is the conditional posterior over the motif location. The fixed point equations are now given by

$$Q_i(z) = ((1 - \lambda)B^0(z) + \lambda m\, n_i(z))/Z_\lambda \qquad (13)$$
$$B(z) = ((1 - \lambda)B^0(z) + \lambda \sum_i n_i(z))/Z_\lambda \qquad (14)$$

where $Z_\lambda = 1 - \lambda + \lambda m \sum_k p_k$. The main difference here is that while the fixed points of $B(\cdot)$ have a similar form, they characterize the unique *minimum* of the objective (by convexity). The background model therefore remains at the worst case choice within the limits allowed by the current allocation $\lambda$.

At $\lambda = 0$, both the motif and the background model reduce to the reference model $B^0$. We subsequently evolve the pair $(Q, B)$ of fixed point solutions from the initial model using the continuation approach. For the resulting model $Q$ to represent a valid motif, we expect that the path passes through critical point(s) ensuring that the motif $Q$ deviates substantially from the competing background. This indeed takes place.

### 5.1 MOTIF FINDING RESULTS

In our experiments, we have used a couple of extensions of the basic approach described above. First, the competing background model has to consider a slightly wider window of basis than the motif. This is reasonable since the background should set the context in which the motif has to differentiate itself. This requires only a minor adjustment to the fixed points[4]. Second, we iteratively find additional motifs by adjusting the prior motif locations $p^0(t)$ for each sequence on the basis of already found motifs. The resulting "search" remains deterministic from the point of view of the algorithm and involves no combinatorial effort.

---
[4] The summation $\sum_i n_i(z)$ in the fixed point equation for $B$ is now over the wider context. The normalization constant has to be adjusted accordingly. Note that the specific minimax optimization problem defined above is no longer strictly speaking valid with this change. The fixed point equations remain meaningful, however, and the continuation approach can be used as before.

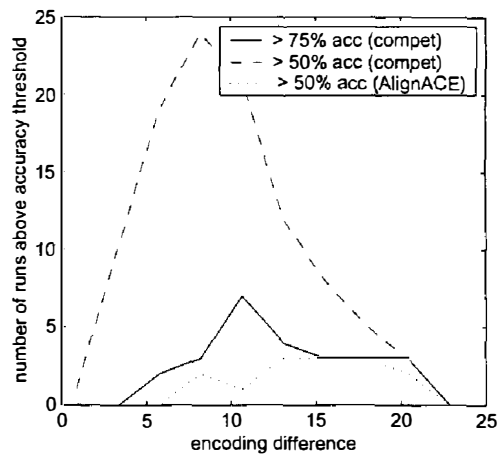

Figure 5: Number of runs above 75% and 50% accuracy against the encoding length. No AlignACE results were above 75%

We have tested the competitive minimax approach against the well-known Gibbs sampling motif software AlignACE (Hughes et al., 2000) on artificially generated data with known motif locations and a significant controlled noise level. We generated sequences from a 5-order Markov background model trained on yeast promoter sequences, and added a single uniformly random motif conserved across all sequences but corrupted by noise at various levels. Problem size was 20 sequences of 200 bases and motifs of length 10. To get comparable results we turned off column sampling in AlignACE (effectively fixing the motif length to the correct one), and set the expected number of sites to 1. We scored only the first motif found by AlignACE and the first motif found by the competition against the motif locations from which we generated the data.

The issue in running a comparison test between such different models is coming up with a fair measure of performance not biased against any of the models. To this end we computed the accuracy of each experimental run as the percentage of sequences in which the found motif was within half motif length of the true location. Sequences with posterior over motif presence smaller than 0.5 were considered errors. Similarly, AlignACE sequences with no motif reported were also errors. We report motif accuracies against the level of noise (*encoding length*) averaged over all sequences, computed in the following manner:

$$\log_2 P_Q(\text{motif}) - \max \log_2 P_Q(\text{other motif locations}) \qquad (15)$$

where the maximum is over all motif locations not overlapping the found motif. The intuition behind encoding length is that sequences with high encoding length have distinguishable motifs. High levels of noise ensure that the encoding length measure is finite.

For a robust comparison we analyze only the number of runs above 75% and 50% accuracy (Figure 5). The competitive motif algorithm consistently finds correct locations more often than AlignACE, even at high noise levels.

## 6 DISCUSSION

We have described a new methodology for addressing a variety of learning tasks that feature a competition between two conflicting sources of information, one of which admits a closed-form (or at least a feasible) solution. The homotopy continuation method follows a continuous path of stationary points of the mixed optimization criterion, by solving a differential equation based on stationarity conditions in $Q(n^3)$ in the number of optimized parameters. By identifying critical points along such paths we can better determine the ideal allocation of the sources.

The method is deterministic and avoids the typical combinatorial complexity of non-convex optimization problems. The key requirement is that the estimation criterion can be reduced to fixed point computations.

The basic methodology admits a number of important extensions from active learning (with the idea of regaining stability) to game theoretic problems more generally.

## APPENDIX

**Theorem 4** *Let $F$ and $G$ be continuous functions on a compact set. There exists a monotonically increasing function $\beta : [0,1] \to [0,\infty)$ such that for every $\lambda \in [0,1]$, $(1-\lambda)F + \lambda G$ and $\{G|F \leq \beta(\lambda)\}$ achieve their minimum at the same time. Moreover, we can choose $\beta(0) = \min F$ and any $\beta(1) \geq \max F$.*

**Proof** $(1-\lambda)F + \lambda G$ is continuous on a compact set thus it achieves its minimum. Let $x_\lambda$ be one of the points achieving it. Moreover, let $x'_\lambda$ be a point achieving the minimum of $\{G|F \leq F(x_\lambda)\}$. Then $F(x'_\lambda) \leq F(x_\lambda)$ and $G(x'_\lambda) \leq G(x_\lambda)$. Therefore $(1-\lambda)F(x'_\lambda) + \lambda G(x'_\lambda) \leq (1-\lambda)F(x_\lambda) + \lambda G(x_\lambda)$, and because $x_\lambda$ is a minimum by definition, we must have equality. If follows that $x'_\lambda$ achieves the minimum of both $\{G|F \leq F(x'_\lambda)\}$ and $(1-\lambda)F + \lambda G$. Therefore we can define $\beta(\lambda) = F(x'_\lambda)$ to ensure that the functions in the statement achieve their minimum at the same time.

It remains to show that $\beta$ is increasing. Let $\lambda_1 \leq \lambda_2$. By definition

$$(1-\lambda_1)F(x'_{\lambda_1}) + \lambda_1 G(x'_{\lambda_1}) \leq (1-\lambda_1)F(x'_{\lambda_2}) + \lambda_1 G(x'_{\lambda_2})$$

$$(1-\lambda_2)F(x'_{\lambda_1}) + \lambda_2 G(x'_{\lambda_1}) \geq (1-\lambda_2)F(x'_{\lambda_2}) + \lambda_2 G(x'_{\lambda_2})$$

Subtracting the second inequality from the first we obtain:

$$(\lambda_2 - \lambda_1)(F(x'_{\lambda_1}) - G(x'_{\lambda_1})) \leq (\lambda_2 - \lambda_1)(F(x'_{\lambda_2}) - G(x'_{\lambda_2}))$$

therefore $F(x'_{\lambda_1}) - G(x'_{\lambda_1}) \leq F(x'_{\lambda_2}) - G(x'_{\lambda_2})$, or equivalently $\beta(\lambda_1) - \beta(\lambda_2) \leq G(x'_{\lambda_1}) - G(x'_{\lambda_2})$.

Assume by contradiction that $\beta(\lambda_1) > \beta(\lambda_2)$. It follows that $G(x'_{\lambda_1}) > G(x'_{\lambda_2})$, and since $F(x'_{\lambda_2}) \leq \beta(\lambda_2) < \beta(\lambda_1)$, $x'_{\lambda_1}$ cannot be a minimum of $\{G|F \leq \beta(\lambda_1)\}$. This is a contradiction. The choice of values of $\beta$ at 0 and 1 is trivial.    □

### Acknowledgements

This work was supported in part by Nippon Telegraph and Telephone Corporation, by ARO MURI grant DAAD 19-00-1-0466, and by NSF ITR grant IIS-0085836.